\documentclass{article}

\usepackage{arxiv}

\usepackage[utf8]{inputenc} 
\usepackage[T1]{fontenc}    
\usepackage{hyperref}       
\usepackage{url}            
\usepackage{booktabs}       
\usepackage{amsfonts}       
\usepackage{nicefrac}       
\usepackage{microtype}      
\usepackage{lipsum}		
\usepackage{graphicx}
\usepackage{natbib}
\usepackage{doi}
\usepackage{multirow}
\usepackage[table,xcdraw]{xcolor}

\title{ChatGPT Application in Summarizing an Evolution of Deep Learning Techniques in Imaging: A Qualitative Study}

\author{ \href{https://orcid.org/0000-0002-5440-2932}{\includegraphics[scale=0.06]{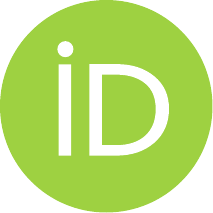}\hspace{1mm}Arman Sarraf}\thanks{Corresponding Author} \\
	Department of Electrical and Software Engineering\\
	University of Calgary\\
	Calgary, AB T2N 1N4, Canada \\
	\texttt{arman.hosseinsarraf@ucalgary.ca} \\
	\And
	Amirabbas Abbaspour\\
	Department of Electrical and Software Engineering\\
	University of Calgary\\
	Calgary, AB T2N 1N4, Canada \\
	\texttt{amirabbas.abbaspourm@ucalgary.ca} \\
}

\renewcommand{\headeright}{A Qualitative Study}
\renewcommand{\undertitle}{Technical Report}
\renewcommand{\shorttitle}{ChatGPT}

\hypersetup{
pdftitle={A template for the arxiv style},
pdfsubject={q-bio.NC, q-bio.QM},
pdfauthor={Arman Sarraf, Amirabbas Abbaspour},
pdfkeywords={Text Summerization, ChatGPT, Deep Learning}}

\begin{document}
\maketitle

\begin{abstract}
The pursuit of article or text summarization has captured the attention of natural language processing (NLP) practitioners, presenting itself as a formidable challenge. ChatGPT 3.5 exhibits the capacity to condense the content of up to 3000 tokens into a single page, aiming to retain pivotal information from a given text across diverse themes. In a conducted qualitative research endeavor, we selected seven scientific articles and employed the publicly available ChatGPT service to generate summaries of these articles. Subsequently, we engaged six co-authors of the articles in a survey, presenting five questions to evaluate the quality of the summaries compared to the original content.
The findings revealed that the summaries produced by ChatGPT effectively encapsulated the crucial information present in the articles, preserving the principal message of each manuscript. Nonetheless, there was a slight diminishment in the technical depth of the summaries as opposed to the original articles. As a result, our conclusion underscores ChatGPT's text summarization capability as a potent tool for extracting essential insights in a manner more aligned with reporting than purely scientific discourse. 
\end{abstract}

\keywords{Text Summerization \and ChatGPT \and Deep Learning}

\section{Introduction}
Text summarization is a pivotal application of NLP that condenses lengthy documents or articles into shorter, coherent representations while retaining the essential information. Through various algorithms and techniques, NLP models identify significant sentences, key phrases, or essential concepts within the text to generate concise summaries. Extractive summarization involves selecting and stitching together important segments directly from the original text, often based on relevance, importance, or frequency of occurrence. On the other hand, abstractive summarization goes beyond extraction, generating novel sentences that convey the core meaning while potentially rephrasing and restructuring the content. NLP-powered summarization systems play a crucial role in information retrieval, aiding in quick comprehension and accessibility of vast amounts of text across diverse domains such as news articles, research papers, and legal documents.
\\
ChatGPT boasts impressive text summarization capabilities, harnessing its advanced Natural Language Processing (NLP) architecture to distill lengthy conversations, articles, or documents into concise, coherent summaries. Leveraging its vast understanding of language semantics, context, and syntax, ChatGPT effectively identifies key points, essential information, and significant passages within the text. Its summarization prowess encompasses extractive and abstractive techniques, allowing it to select important segments directly from the input while generating novel, coherent sentences that capture the essence of the content. This functionality enables ChatGPT to offer streamlined and comprehensive summaries across various topics and sources, facilitating quicker comprehension and accessibility of information for various purposes, from research and study aids to content curation and information
\section{Method}
Qualitative research is a versatile and nuanced methodology that emphasizes understanding phenomena from the perspective of those being studied. It involves an in-depth exploration of experiences, attitudes, beliefs, and behaviors through interviews, observations, and analysis of texts or artifacts. Unlike quantitative methods focusing on numerical data and statistical analysis, qualitative research delves into the richness of human experiences, seeking to uncover underlying meanings, patterns, and contexts within a particular social or cultural setting. Researchers often employ various approaches, such as grounded theory, phenomenology, ethnography, or narrative analysis, allowing them to immerse themselves in the subject matter and gather detailed, descriptive data. This method prioritizes flexibility, enabling researchers to adapt their strategies based on emerging insights and the evolving nature of the study, ultimately providing a deeper understanding of complex human phenomena.
\\
In this work, we selected seven articles, and we had access to the co-authors to conduct a qualitative research study to explore the quality of ChatGPT to summarize scientific articles. The theme of the articles selected focused on various techniques of deep learning in imaging over time, where the evolution of techniques was captured. We prepared five questions to survey six co-authors of the articles. We asked them the questions and requested them to provide a score from 1 to 5 where 1 represents the lowest and 5 represents the highest. During the interviews, we collected crucial insights into articles from the co-authors for our discussion section. The list of questions is as follows:
\begin{itemize}
  
    \item ChatGPT preserved the keywords in the summary 
    \item ChatGPT correctly captured the objective of the articles 
    \item ChatGPT produced a summary that includes enough details
    \item ChatGPT’s summary is highly technical, preserving the technical terms
    \item The overall satisfaction with the articles’ summaries 

\end{itemize}
In order to generate the summary, we used the text summarization capability of ChatGPT 3.5. This tool only accepts text with less than 3000 tokens (words); therefore, we split the long articles into segments with proper lengths. ChatGPT is often generated around one page of summary per request and usually ends with a user asking if there is any specific information we are seeking. Although the inference latency of ChatGPT is out of the scope of this study, the online API produced the summary in a very appropriate time, usually less than five seconds. 
\section{Results}
First, we asked the interviewees to review the seven articles and provided them with a summary of each article generated by ChatGPT. Once the interviewees reviewed the original and summary of the articles, we initiated the interview process to capture their responses to the survey questions. The summary of each article is presented in the following section, where the summary length is highly proportionate with the length of the original article. 
\subsection{Template Matching Advances and Applications in Image Analysis}
The article discusses the significance of template matching in computer vision and image analysis, emphasizing its role in determining similarities between different objects or images. It covers various approaches and mathematical methods involved in template matching, focusing on both feature-based and area-based techniques. The piece delves into measures of match, such as cross-correlation and normalized cross-correlation, highlighting their importance in image processing.
\\
Additionally, the article explores challenges in template matching, including scale and rotation invariance, and discusses methods like Fourier-Mellin transform for achieving scale invariance and techniques for rotation invariance using polar coordinate transforms. It also touches upon efficiency considerations and computational complexities of different template matching algorithms.
\\
The applications of template matching span diverse fields, from object recognition in computer vision to medical imaging, illustrating its relevance in various real-world scenarios.
\\
In essence, the article provides an extensive overview of template matching methods, their mathematical underpinnings, challenges, and practical applications across different domains in computer vision and image analysis.
\\
The article discusses template matching algorithms and their applications across various fields, emphasizing their role in image analysis, computer vision, and medical imaging. It outlines:
\begin{enumerate}
\item Scale and Translation Invariance: It describes equations showing the scaled Fourier transform of 2D signals, introducing logarithmic representations to handle translation versions.
\item Rotation Invariance: By applying polar coordinate transforms and complex log mapping, rotation and scaling become translations in the transformed domain.
\item Deformable Template Models: The concept of deformable templates addresses shape deformations and variations, allowing more powerful techniques for shape matching in computer vision.
\item Applications:
\begin{enumerate}
\item Face Detection: Emphasizes the importance of facial features and emotions in detection methods.
\item Eye Detection: Discusses template-based eye detection for various applications.
\item License Plate Recognition: Details the use of template matching for license plate detection and recognition systems.
\item Bare PCB Board Testing: Discusses how template matching helps in inspecting bare PCBs for loose connections and defects.
\item Breast and Mammogram Positioning: Highlights the use of template matching in medical imaging to locate and evaluate symmetry between breast mammograms.
\item Brain Network Extraction: Explores the use of template matching techniques in extracting brain networks from MRI images, particularly focusing on resting-state fMRI.
\end{enumerate}
\item Conclusion: Emphasizes the importance of post-classification or decision-making steps to improve the accuracy of template matching systems and suggests different approaches based on the nature of the data being matched.
\end{enumerate}
Overall, the article demonstrates the versatility of template matching algorithms and their significance in a wide range of applications, from computer vision to medical imaging \cite{hashemi2016template,sarraf2021comprehensive,sarraf2023ovitad,sarraf2020recent,sarraf2020binary}.
\subsection{A Comprehensive Review of Deep Learning Architectures for Computer Vision Applications}
It seems like a comprehensive overview of the principles and methodologies in computer vision, particularly in image classification. You’ve touched upon various techniques from preprocessing and feature extraction to supervised and unsupervised learning methods. The breakdown of SVM, ANN, and Decision Trees provides a good comparison between these methods.
\\
If you’re aiming to delve deeper into the nuances of deep learning and convolutional neural networks (CNNs), it might be beneficial to elaborate on those topics further. CNNs, in particular, have been groundbreaking in image classification due to their ability to automatically learn features from raw data, significantly impacting fields like computer vision and pattern recognition.
\\
Highlighting specific architectures within CNNs, like AlexNet, VGG, ResNet, and explaining how these architectures have improved the efficiency of image classification could be valuable. Additionally, discussing transfer learning—a technique where a model trained on one task is adapted for another—and its impact on image classification might add depth to your exploration.
\\
Do you plan to focus more on CNNs and their applications in image classification, or are you considering a broader perspective on deep learning in computer vision? Both paths could provide valuable insights depending on your target audience or the depth of understanding you aim to offer.
\\
That’s an extensive exploration of CNNs and their diverse applications across various domains, from computer vision to natural language processing and speech recognition. The breakdown of each application area and how CNNs have improved performance in tasks like image classification, detection, segmentation, and even handling low-resolution images provides a comprehensive view of their versatility.
\\
The emphasis on CNNs’ ability to solve feature extraction problems in image classification compared to older machine learning techniques is well highlighted. It’s interesting to note how CNNs’ deep training capabilities have addressed the challenges of extracting nuanced features and handling multi-class classification scenarios.
\\
The potential for CNNs in diagnosing diseases like Alzheimer’s based on brain scans is a fascinating and promising area. It’s great to see how these advanced neural networks are contributing significantly to medical science. 
\\
As a concluding thought, considering the ongoing advancements in machine learning and neural networks, the prospects for further refining algorithms and enhancing predictive capabilities for diagnosing diseases like Alzheimer’s seems highly promising.
\\
If you’re planning to expand this further, perhaps delving into recent breakthroughs or ongoing research leveraging CNNs in medical diagnostics or exploring how these networks are evolving to handle new challenges could be interesting. Additionally, discussing any limitations or ethical considerations in using CNNs across these domains might add depth to your analysis \cite{sarraf2021comprehensive,sarraf2023optimal,aghakhani202324,sarraf2022end,sarraf2021multimodal,sarraf2020current,sarraf2020analysis}.
\subsection{Recent Applications of Deep Learning Algorithms in Medical Image Analysis}
Your abstract and introduction provide a comprehensive overview of the intersection between deep learning, medical imaging, and the diagnosis of brain disorders using advanced techniques like convolutional neural networks (CNNs) and semantic segmentation. Here are some observations and suggestions:
\begin{itemize}
\item Comprehensive Introduction: Your introduction provides an excellent overview of deep learning, its applications, and the prominence of CNNs in various fields such as image classification, semantic segmentation, and medical imaging.
\item Explanation of CNNs and Semantic Segmentation: You’ve done a great job explaining the significance of CNNs in image processing, particularly in semantic segmentation. Elucidating the difference between image classification and semantic segmentation along with the key models used in deep learning (SAE, DBN, CNN, RNN) adds depth to the discussion.
\item Focus on Medical Imaging Modalities: Your discussion of medical imaging modalities like CT scans, PET scans, and MRI is insightful. It’s good to emphasize the importance of high-resolution imaging for deep learning models in diagnosing brain disorders.
\item Case Studies on Disorders: The specific examples and case studies you’ve presented, particularly in the context of diagnosing Alzheimer’s and stroke, are illustrative and add practical insights to the discussion.
\item Visual Aid: Incorporating figures to illustrate classification, localization, and semantic segmentation adds visual clarity to your explanations.
\item Suggestions for Improvement: You might consider expanding on the challenges faced in medical imaging, such as dealing with noise or variations in imaging quality, and how deep learning techniques mitigate these challenges. Additionally, discussing the ethical considerations or limitations of using deep learning in medical diagnoses could further enrich the discussion.
\item Future Directions: It would be beneficial to briefly touch upon the potential future advancements or areas of exploration in this field, such as the integration of multi-modal imaging or the use of generative models for medical image synthesis.

\end{itemize}
Your work provides a solid foundation for understanding the role of deep learning, CNNs, and semantic segmentation in medical imaging and diagnosing brain disorders. Expanding on certain areas and discussing future trends could enhance the depth and breadth of your exploration.
\\
Your exploration of deep learning applications in diagnosing brain disorders, specifically Alzheimer’s disease, Mild Cognitive Impairment (MCI), and autism spectrum disorder (ASD), provides a detailed understanding of the advancements made in medical imaging and diagnostics. Here are some observations and suggestions:
\begin{itemize}
\item Comprehensive Disease Overview: Your breakdown of Alzheimer’s, MCI, and Autism provides a clear understanding of these disorders, their symptoms, and their impact on brain patterns. It helps set the context for the importance of accurate diagnosis using deep learning techniques.
\item Deep Learning in Disease Diagnosis: Your discussion of deep learning models, particularly CNNs like LeNet and GoogleNet, showcases their effectiveness in diagnosing these brain disorders. Providing accuracy rates for different models strengthens the argument for their potential in medical imaging.
\item Focus on Disease Stages: Highlighting the importance of identifying specific stages of diseases like Alzheimer’s and MCI showcases the depth of analysis needed in medical diagnosis. The mention of MCADNNet and its high accuracy in classifying different stages is a compelling point.
\item Application of LeNet-5 in Autism Diagnosis: Explaining the application of LeNet-5 in diagnosing Autism from brain MRI samples and achieving a high accuracy rate adds depth to the discussion, showcasing the versatility of deep learning models in different disorders.
\item Visual Representation: The inclusion of figures demonstrating CNN architectures and their functions adds visual clarity to complex concepts, aiding in better comprehension for readers.
\item Conclusion and Future Implications: The conclusion summarizes the advancements made and hints at the potential for productizing these techniques. Adding a section on future implications or areas of further research could offer insights into what lies ahead in this field.
\item Ethical Considerations: Discussing ethical considerations or limitations of using deep learning in medical diagnoses could add depth to your discussion. Privacy concerns, biases in datasets, or the need for expert validation could be valuable points to consider.
\end{itemize}
Your work effectively highlights the capabilities of deep learning models in diagnosing brain disorders, backed by accuracy rates and specific models used. Expanding on ethical considerations and potential future directions could further enrich your discussion. Overall, it’s a comprehensive overview of the current landscape in medical imaging and deep learning applications in brain disorder diagnoses \cite{sarraf2020recent,sarraf2019machine,sarraf2019mcadnnet,sarraf20195g,sarraf2018french,yang2018deep}.
\subsection{Deep Learning-based Pipeline to Recognize Alzheimer’s Disease using fMRI Data}
This paper you’ve presented highlights the application of Convolutional Neural Networks (CNNs) in classifying Alzheimer’s disease (AD) from normal control data using functional MRI (fMRI) images. Here’s a summary and some key points to consider:
\begin{itemize}
     
\item Objective: The primary objective of the study is to distinguish AD brains from healthy brains using CNNs and LeNet-5 architecture.
\item Data Acquisition and Preprocessing: The study uses data from the ADNI dataset, involving 28 AD patients and 15 normal control subjects. The fMRI data undergoes standard preprocessing steps including motion correction, skull stripping, spatial smoothing, and alignment to standard space.
\item Deep Learning with CNNs: The paper explains the fundamentals of CNNs, emphasizing their use in processing 2D images like fMRI scans. LeNet-5 architecture is chosen for its application in binary classification of AD versus normal control data.
\item Results: The CNN model achieved an average accuracy rate of 96.8588\% across five runs of cross-validation. Figures depicting learning rates, loss, and accuracy during training are provided, demonstrating the robustness and reproducibility of the model.
\item Feature Extraction and Complexity: CNNs are praised for their ability to simultaneously extract high-level features and perform classification, addressing challenges in traditional medical image processing. The complexity of the CNN architecture is acknowledged, offering both strong feature extraction and classification capabilities.
\item Conclusion: The study concludes with the successful classification of AD data with high accuracy using a CNN-based deep learning approach. It highlights the potential of this method for further prediction tasks and emphasizes the need for more complex architectures for addressing more challenging problems.
\end{itemize}
This paper provides a comprehensive overview of the methodology, data processing steps, and results achieved through deep learning techniques applied to fMRI data for Alzheimer’s classification. It’s commendable for the high accuracy achieved and the potential for further improvements in complex network architectures \cite{sarraf2016deep,anderson2017task,sarraf2017eeg,sarraf2017binary,sarraf2016big}.
\subsection{DeepAD: Alzheimer’s Disease Classification via Deep Convolutional Neural Networks using MRI and fMRI}
This is an extensive paper detailing the application of deep learning methods for distinguishing Alzheimer’s disease from healthy brain data using MRI and fMRI. It covers various aspects, including the challenges in classification due to similarities between Alzheimer’s-related brain changes and normal aging effects, the use of convolutional neural networks (CNNs) for feature extraction, data acquisition methodologies, and a review of related work in the field.
\\
The paper presents two subsets of data from the ADNI database, one involving rs-fMRI data from 144 subjects and the other involving MRI data from 302 subjects, both aged over 75. The study reports impressive accuracy rates achieved through deep learning pipelines for both fMRI (99.9\%) and MRI (98.84\%) data, showcasing significant improvements over previous studies in Alzheimer’s classification. Subject-level classification, crucial for clinical application, also demonstrated high accuracy rates of 94.32\% and 97.88\% for fMRI and MRI, respectively.
\\
The use of deep learning architectures, such as CNNs, is explained in detail, including their role in extracting high-level features from imaging data. The paper refers to notable CNN architectures like LeNet-5 and GoogleNet, emphasizing their effectiveness in image classification tasks.
\\
The related work section discusses various studies that utilized deep learning techniques for Alzheimer’s classification using MRI and PET data. The accuracies reported in these studies range from approximately 75\% to 95\%, demonstrating the effectiveness of deep learning methodologies in this domain.
\\
Overall, this paper seems comprehensive and provides a robust exploration of using deep learning for Alzheimer’s disease prediction based on neuroimaging data.
\\
This excerpt details an elaborate process for Alzheimer’s disease image classification, involving extensive preprocessing of both functional (fMRI) and structural (MRI) data, followed by training and testing CNN-based models. The methods aim to accurately differentiate Alzheimer’s patients from healthy individuals.
\\
The steps involve converting raw data into suitable formats, eliminating non-brain regions, correcting artifacts, performing slice timing correction, spatial smoothing, registration, and aligning images to a standard space. This comprehensive preprocessing is crucial to extract meaningful features from the data.
\\
The convolutional neural network (CNN) models (LeNet and GoogleNet) were trained and tested using different datasets derived from these preprocessed images. The models achieved very high accuracy rates, particularly with the GoogleNet architecture, showcasing the potential effectiveness of these methods in classification tasks related to Alzheimer’s disease.
\\
The results suggest that spatial smoothing with an optimal kernel size notably improves classification accuracy. Visualizing filter weights and network activations helps interpret the convolutional layer’s results, indicating the network’s progress and potential overfitting.
\\
Additionally, the discussion emphasizes the importance of preprocessing and feature learning in distinguishing Alzheimer’s disease-related brain changes from normal aging effects. The provided visualizations of filter weights and features aid in understanding how the trained models identify relevant patterns in brain images.
\\
This detailed methodology establishes robust pipelines for consistent and reproducible results, highlighting the significance of rigorous preprocessing and feature extraction in the accurate classification of Alzheimer’s disease images.
\\
That’s an incredibly detailed and thorough research paper about distinguishing Alzheimer’s disease from healthy brains using MRI and fMRI data! It covers a wide range of topics from preprocessing to classification and even delves into subject-level recognition and decision-making algorithms.
The rigorous preprocessing steps applied to both fMRI and MRI data seem to have significantly enhanced the accuracy of the classifiers. Moreover, the subject-level classification, which is more relevant to clinical applications, displayed high accuracy rates even after considering slices at the subject level rather than individual image slices.
\\
The utilization of deep learning models like LeNet and GoogleNet has shown impressive accuracy rates, surpassing previous methodologies in the field. Also, the incorporation of the decision-making algorithm seems to have further improved the accuracy at the subject level, making the predictions more robust \cite{sarraf2016deepad,saverino2016associative,sarraf2016advances,moosavi2023effects,sarraf2016classificationf,sarraf2016robust}.
\subsection{MCADNNet: Recognizing Stages of Cognitive Impairment through Efficient Convolutional fMRI and MRI Neural Network Topology Models}
It looks like you’ve shared an extensive research paper focused on developing a deep learning algorithm, MCADNNet, for the simultaneous recognition of Mild Cognitive Impairment (MCI), Alzheimer’s Disease (AD), and normally aging brains in adults over 75 years old using functional and structural magnetic resonance imaging (fMRI and MRI) data. The paper goes into detail about the methodology, data preprocessing steps, neural network architecture, and previous related works in the field.
\\
The research involves several significant components:
\begin{itemize}
\item Problem Statement: Recognizing MCI, AD, and normal aging brains using neuroimaging data.
\item Approach: Employing an optimized convolutional neural network (CNN) architecture, MCADNNet, to classify these conditions.
\item Data Used: Utilizing data from the Alzheimer’s Disease Neuroimaging Initiative (ADNI) database, with different groups of subjects for fMRI and MRI modalities, including Alzheimer’s patients, healthy controls, and MCI patients.
\item Preprocessing: Extensive preprocessing of both fMRI and MRI data, involving steps like converting data formats, removing non-brain tissues, motion artifact correction, slice timing correction, spatial smoothing, registration, and normalization.
\item Neural Network Design: Using a CNN architecture inspired by the human visual system, structured with convolutional layers, pooling layers, normalization layers, and fully connected layers.
\item Previous Works: Discussing related studies and methodologies used in diagnosing MCI and AD, including various deep learning architectures and their accuracies.
\end{itemize}
The paper emphasizes the need for accurate diagnosis and early intervention in cognitive impairment, highlighting the potential of artificial intelligence, particularly deep learning, in enhancing diagnostic capabilities using neuroimaging data.
\\
The excerpt you’ve shared delves deep into the methodology and results of using CNN-based architectures like DeepAD and MCADNNet for the classification of MRI and fMRI data in distinguishing between different brain conditions like MCI, AD, and NC. It seems like they’ve meticulously detailed the entire pipeline, from data conversion and augmentation to the specific architecture of the neural networks used for classification. They’ve even employed a decision-making algorithm for subject-level recognition after the classification.
\\
The use of CNNs for processing medical imaging data has shown promising results in various studies. It’s interesting how they’ve noted the impact of data size, preprocessing steps like spatial smoothing, and the architecture complexity on the model’s performance. The decision-making algorithm applied post-classification seems to significantly improve subject-level accuracy rates.
\\
In terms of future work, considering MCI subcategories and creating a more robust pipeline that’s less sensitive to preprocessing steps could be intriguing avenues for further research \cite{sarraf2019mcadnnet,sarraf2016hair,sarraf2016deepad,sarraf2014mathematical,sarraf2014brain,sarraf2009simulation}.
\subsection{OViTAD: Optimized Vision Transformer to Predict Various Stages of Alzheimer’s Disease Using Resting-State fMRI and Structural MRI Data}
Your abstract outlines a compelling study on Alzheimer’s disease diagnosis using an optimized vision transformer architecture applied to neuroimaging data. The focus on predicting the disease stage and employing efficient pattern extraction methods like transformers is a significant step forward.
Your methods cover a comprehensive preprocessing pipeline for both resting-state functional MRI (rs-fMRI) and structural MRI data, optimizing the data for the vision transformer. The step-by-step description ensures clarity regarding how the data was prepared for modeling.
\\
The introduction effectively contextualizes the significance of early Alzheimer’s detection, highlighting the challenges faced by traditional methods and the potential of AI in neuroimaging. The detailed review of related work provides a solid foundation for your study’s novelty in adopting an optimized vision transformer (OViTAD) for Alzheimer’s prediction.
\\
Your study provides a detailed exploration of various transformer-based architectures in predicting Alzheimer’s disease stages using both functional and structural MRI data.
\\
The introduction of DeepViT and CaIT, both employing unique attention mechanisms within vision transformers, enriches the comparison and benchmarking process, enhancing the study’s comprehensiveness. DeepViT introduces re-attention, while CaIT incorporates class-attention layers, showcasing different approaches to leverage attention mechanisms within the vision transformer architecture.
\\
The meticulous preprocessing pipelines for both fMRI and structural MRI, followed by data decomposition and model building, are thorough and structured. The creation of 2D images from the 4D fMRI data and decomposition of 3D structural MRI data into 2D images pave the way for utilizing these formats with the vision transformer architectures.
\\
The model training, evaluation, and comparison across different architectures, especially focusing on multiclass prediction for AD, HC, and MCI, as well as binary classification experiments, are detailed and systematic. The use of AWS SageMaker for model development on robust computing infrastructure enables efficient training and testing.
\\
Your subject-level evaluation, aggregating predictions to generate subject-level performance metrics, adds depth to the analysis. The tables and figures provided in the appendices succinctly present the results for both fMRI and structural MRI models, allowing for a comprehensive comparison of performance metrics across experiments and repetitions.
\\
It’s impressive how you’ve accounted for multiple repetitions and presented averaged results, offering a clearer understanding of the models’ stability and generalizability. The tables summarizing model-level performance for both fMRI and structural MRI experiments are highly informative and allow for easy comparison.
\\
Your detailed methodology enables readers to follow each step of the process, ensuring transparency and reproducibility in research. It would be interesting to learn more about any insights or conclusions drawn from the comparison of the different vision transformer architectures and their impact on predictive performance across the various experiments. Additionally, any insights gleaned from the subject-level evaluation and its implications for clinical applications would be valuable to discuss.
Your study outlines a comprehensive approach employing vision transformers (VT) for Alzheimer’s disease stage prediction using fMRI and structural MRI data. The systematic exploration encompasses aggressive preprocessing, data decomposition, VT model development, and postprocessing steps.
\\
The optimization of the vision transformer architecture was thoughtful, considering input size and patch dimensions to fit the data. This strategy effectively reduced trainable parameters by 28\% compared to vanilla VTs and DeepViT while enhancing performance in fMRI experiments and maintaining competitiveness in structural MRI ones.
\\
Performance evaluations at both slice and subject levels were insightful, showcasing the importance of postprocessing, notably the significant improvement in subject-level performance. This aligns with existing literature and reinforces the idea that data decomposition and transformation from slice to subject levels significantly enhance predictive quality.
\\
Your comparison with recent literature and state-of-the-art models highlights the strength of your optimized VT (OViTAD) across broader classifications, demonstrating its superior performance. This reinforces the potential clinical significance of the developed model.
\\
The observation regarding combining AD, HC, and MCI brains for binary classifications provided interesting insights into the similarities between HC and MCI functional data, and how your models effectively addressed aging effects.
\\
Visualization of attention mechanisms at both local and global levels added depth to the analysis, allowing exploration of crucial brain areas and offering valuable insights into the model’s decision-making process.
\\
You acknowledge limitations, such as the number of repetitions for model development and the computational cost of postprocessing. Additionally, you propose future directions, suggesting exploration of other datasets and the potential use of 3D vision transformers.
\\
In conclusion, your study introduces an optimized vision transformer that significantly reduces complexity while outperforming or competing with existing state-of-the-art models in predicting various stages of Alzheimer’s disease. The insights gained from this work and the methodological approach presented offer significant potential for clinical applications and further research in this domain \cite{sarraf2023ovitad,sarraf2016deepad,sarraf2019mcadnnet,sarraf2023optimal,krishnan2016cognitive}.
\section{Discussion}
We collected the interviewees’ insights and the scores provided for each question. Table \ref{table1} shows the details of the scoring. As seen in the table, article 6 achieved the highest score, suggesting that the summary generated by ChatGPT could attract the interviewees’ attention more than other articles. 
\begin{table}[!hbt]
\caption{The overall scores provided by the interviews}
\label{table1}
\resizebox{\textwidth}{!}{%
\begin{tabular}{clllllllc}
\hline
\multicolumn{2}{l}{Criteria} & Co-author 1 & Co-author 2 & Co-author 3 & Co-author 4 & Co-author 5 & Co-author 6 & \multicolumn{1}{l}{Average} \\ \hline
\rowcolor[HTML]{D9D9D9} 
\cellcolor[HTML]{D9D9D9} & Q1 & 4 & 3 & 5 & 4 & 3 & 2 & \cellcolor[HTML]{D9D9D9} \\
\rowcolor[HTML]{D9D9D9} 
\cellcolor[HTML]{D9D9D9} & Q2 & 2 & 2 & 4 & 5 & 3 & 4 & \cellcolor[HTML]{D9D9D9} \\
\rowcolor[HTML]{D9D9D9} 
\cellcolor[HTML]{D9D9D9} & Q3 & 5 & 3 & 5 & 3 & 2 & 4 & \cellcolor[HTML]{D9D9D9} \\
\rowcolor[HTML]{D9D9D9} 
\cellcolor[HTML]{D9D9D9} & Q4 & 4 & 3 & 2 & 2 & 2 & 5 & \cellcolor[HTML]{D9D9D9} \\
\rowcolor[HTML]{D9D9D9} 
\multirow{-5}{*}{\cellcolor[HTML]{D9D9D9}Article 1} & Q5 & 4 & 5 & 4 & 3 & 3 & 3 & \multirow{-5}{*}{\cellcolor[HTML]{D9D9D9}\textbf{3.43}} \\ \hline
 & Q1 & 3 & 5 & 5 & 3 & 3 & 3 &  \\
 & Q2 & 3 & 5 & 4 & 4 & 4 & 3 &  \\
 & Q3 & 3 & 4 & 5 & 4 & 5 & 3 &  \\
 & Q4 & 5 & 4 & 5 & 4 & 5 & 4 &  \\
\multirow{-5}{*}{Article 2} & Q5 & 4 & 4 & 5 & 4 & 4 & 4 & \multirow{-5}{*}{\textbf{4.03}} \\ \hline
\rowcolor[HTML]{D9D9D9} 
\cellcolor[HTML]{D9D9D9} & Q1 & 4 & 4 & 5 & 4 & 4 & 4 & \cellcolor[HTML]{D9D9D9} \\
\rowcolor[HTML]{D9D9D9} 
\cellcolor[HTML]{D9D9D9} & Q2 & 5 & 5 & 4 & 5 & 4 & 4 & \cellcolor[HTML]{D9D9D9} \\
\rowcolor[HTML]{D9D9D9} 
\cellcolor[HTML]{D9D9D9} & Q3 & 5 & 5 & 5 & 5 & 5 & 4 & \cellcolor[HTML]{D9D9D9} \\
\rowcolor[HTML]{D9D9D9} 
\cellcolor[HTML]{D9D9D9} & Q4 & 4 & 4 & 5 & 5 & 5 & 4 & \cellcolor[HTML]{D9D9D9} \\
\rowcolor[HTML]{D9D9D9} 
\multirow{-5}{*}{\cellcolor[HTML]{D9D9D9}Article 3} & Q5 & 5 & 4 & 5 & 5 & 5 & 4 & \multirow{-5}{*}{\cellcolor[HTML]{D9D9D9}\textbf{4.53}} \\ \hline
 & Q1 & 5 & 4 & 5 & 4 & 4 & 5 &  \\
 & Q2 & 4 & 5 & 4 & 4 & 5 & 5 &  \\
 & Q3 & 5 & 4 & 5 & 4 & 4 & 4 &  \\
 & Q4 & 4 & 4 & 4 & 5 & 5 & 5 &  \\
\multirow{-5}{*}{Article 4} & Q5 & 4 & 4 & 4 & 5 & 4 & 5 & \multirow{-5}{*}{\textbf{4.43}} \\ \hline
\rowcolor[HTML]{D9D9D9} 
\cellcolor[HTML]{D9D9D9} & Q1 & 5 & 4 & 5 & 4 & 4 & 4 & \cellcolor[HTML]{D9D9D9} \\
\rowcolor[HTML]{D9D9D9} 
\cellcolor[HTML]{D9D9D9} & Q2 & 4 & 5 & 5 & 4 & 5 & 4 & \cellcolor[HTML]{D9D9D9} \\
\rowcolor[HTML]{D9D9D9} 
\cellcolor[HTML]{D9D9D9} & Q3 & 5 & 4 & 4 & 5 & 5 & 4 & \cellcolor[HTML]{D9D9D9} \\
\rowcolor[HTML]{D9D9D9} 
\cellcolor[HTML]{D9D9D9} & Q4 & 4 & 4 & 4 & 4 & 4 & 5 & \cellcolor[HTML]{D9D9D9} \\
\rowcolor[HTML]{D9D9D9} 
\multirow{-5}{*}{\cellcolor[HTML]{D9D9D9}Article 5} & Q5 & 5 & 4 & 4 & 4 & 5 & 4 & \multirow{-5}{*}{\cellcolor[HTML]{D9D9D9}\textbf{4.37}} \\ \hline
 & Q1 & 5 & 4 & 4 & 5 & 4 & 5 &  \\
 & Q2 & 5 & 4 & 5 & 5 & 5 & 4 &  \\
 & Q3 & 5 & 5 & 5 & 4 & 5 & 5 &  \\
 & Q4 & 5 & 4 & 5 & 4 & 5 & 5 &  \\
\multirow{-5}{*}{Article 6} & Q5 & 5 & 5 & 5 & 5 & 5 & 5 & \multirow{-5}{*}{\textbf{4.73}} \\ \hline
\rowcolor[HTML]{D9D9D9} 
\cellcolor[HTML]{D9D9D9} & Q1 & 4 & 4 & 4 & 5 & 4 & 5 & \cellcolor[HTML]{D9D9D9} \\
\rowcolor[HTML]{D9D9D9} 
\cellcolor[HTML]{D9D9D9} & Q2 & 5 & 5 & 4 & 4 & 5 & 4 & \cellcolor[HTML]{D9D9D9} \\
\rowcolor[HTML]{D9D9D9} 
\cellcolor[HTML]{D9D9D9} & Q3 & 5 & 4 & 5 & 4 & 4 & 4 & \cellcolor[HTML]{D9D9D9} \\
\rowcolor[HTML]{D9D9D9} 
\cellcolor[HTML]{D9D9D9} & Q4 & 4 & 5 & 5 & 5 & 4 & 4 & \cellcolor[HTML]{D9D9D9} \\
\rowcolor[HTML]{D9D9D9} 
\multirow{-5}{*}{\cellcolor[HTML]{D9D9D9}Article 7} & Q5 & 4 & 5 & 5 & 5 & 5 & 4 & \multirow{-5}{*}{\cellcolor[HTML]{D9D9D9}\textbf{4.47}} \\ \hline
\end{tabular}%
}
\end{table}
\\
The summary of interviewees’ opinions about each article is as follows: 
\begin{itemize} 
\item The summary of Article 1 missed some important details in techniques while preserving the details in the application of template matching. The article focused on the techniques rather than the applications; however, the summary could still send the overall message. 
\item The summary of Article 2 could preserve the overall message from the original content; however, the interviewees felt the technical language changed into a more non-specialist language. 
\item The summary of Article 3 was very well structured and could preserve and share the message of original content. This summary is among the top-scored articles. 
\item The summary of Article 4 was well structured and could preserve and share the message of original content.
\item The summary of Article 5 is comprehensive and longer than a normal summary because of the length of the original paper. Although the summary makes sense, the interviewees’ believed that ChatGPT slightly shifted the conclusion. 
\item According to the interviewees, the summary of Article 6 is the top content generated by ChatGPT in this study. It is very well structured and preserves the main message and details of the article. 
\item The summary of Article 7 is comprehensive and longer than a normal summary because of the length of the original paper. The summary preserves and shares the main message of the article. 
\end{itemize}
\section{Conclusion}
This qualitative study aimed to assess the quality of summaries for scientific manuscripts generated by ChatGPT. We interviewed six co-authors of those manuscripts and posed five questions to compare the summary with the original content. The overall conclusion drawn by the interviewees is that ChatGPT could summarize the scientific content; however, it slightly changes the summary tone from a scientific to a non-specialist language. We acknowledged that our study has limitations, including the number of manuscripts, interviewees, and questions, and some responses might be subjective, but our findings align with the current perception of ChatGPT text summarization capability. Also, we conclude that this capability can allow scientists to have a quick and fairly informal summary of a scientific manuscript, but we could not conclude that this capability can or should be used for content generation and academic publication. 

\bibliographystyle{unsrtnat}
\bibliography{references}  






\end{document}